\documentclass{article}
\usepackage[dblblindworkshop, final]{neurips_2026}
\workshoptitle{TS-LIMITS}
\usepackage{amsmath}
\usepackage{graphicx}
\usepackage{natbib}
 \usepackage{amssymb}
\usepackage[utf8]{inputenc}
\usepackage[T1]{fontenc}
\usepackage{hyperref}
\usepackage{url}
\usepackage{booktabs}
\usepackage{amsfonts}
\usepackage{nicefrac}
\usepackage{microtype}
\usepackage{xcolor}
\usepackage{soul}
\usepackage{threeparttable}
\definecolor{darkgreen}{RGB}{0,100,0}
\newcommand{\illy}[1]{\textcolor{black}{#1}}

\title{Energy and Latency First: How Deployment Constraints Reverse Time-Series Model Selection}

\author{%
\textbf{Shruti Bothe}$^{*}$, \textbf{Illyyne Saffar}$^{\P}$, \textbf{Ayush K. Varshney}$^{\S}$ \\[0.5em]
Ericsson Research\\
Santa Clara, USA$^{*}$ \quad Massy, France$^{\P}$ \quad Kista, Sweden $^{\S}$ \\[0.5em]
\begin{tabular}{c}
\texttt{\{shruti.bothe, illyyne.saffar, ayush.kumar.varshney\}@ericsson.com}
\end{tabular}
}
\begin{document}
\maketitle

\begin{abstract}
AI models, and time-series models in particular are typically ranked by accuracy and deployed under constraints absent from that ranking. 
In this paper, we study a real setting where these constraints are explicit and measurable: handover prediction in cellular networks, where inference runs on the User Equipment once per 40\,ms measurement period and must complete before the network acts.
The timing bound is hard: a late prediction is superseded by the measurement it was predicting, while energy is a cumulative budget spent for the life of the device. Neither appears in the objective these models are selected on. We evaluate eight architectures under a protocol that screens for deadline feasibility at the tail rather than the mean, then ranks the feasible models on measured inference energy, counting the feature computation that precedes the forward pass and is usually omitted. Gradient-boosted trees on handcrafted temporal features are the most accurate model (F1=0.974) and among the cheapest to run, while the neural architecture an accuracy-only benchmark would rank first (CNN-1D, F1=0.914) costs over 3$\times$ more energy per inference under a realistic duty-cycled deployment pattern. We give three conditions under which this outcome should be expected, so the finding applies beyond telecommunications. Accuracy-only benchmarking does not merely under-report deployment cost. It can rank first a model that cannot be deployed.
\end{abstract}

\section{Introduction}
Time-series models are ranked by accuracy, then deployed in conditions the benchmark never tested: offline, in batch, on hardware unlike the target device. Yet the applications that need time-series prediction most, such as sensors, vehicles and other edge devices, must decide locally, fast, and within a tight energy budget. Where deployment cost is reported at all, it is approximated by parameter count or FLOPs, which correlate poorly with real device energy~\cite{energy2023flops}, or by mean latency, which says nothing about whether every inference meets a deadline. This paper studies a scenario where both constraints can be measured directly.
  
A mobile user expects uninterrupted connectivity. A cellular network is divided into cells, each served by a base station; at any moment the user equipment (UE) is connected to one serving cell. As the UE moves, its serving signal weakens while a neighbouring cell's may strengthen, and the network must transfer the connection without dropping the session, a procedure called a handover. The decision follows the 3GPP A3 rule~\cite{3gpp38331}: a neighbour becomes a candidate once its RSRP (Reference Signal Received Power, reported every 40\,ms) exceeds the serving cell's by a fixed margin, sustained for a Time-to-Trigger window. Our task is to predict whether this condition will be met within a short horizon, so the network can prepare the handover proactively rather than react after the threshold is crossed. In 5G and beyond the reactive rule is reaching its limits: cells are smaller and frequencies higher, so handovers grow more frequent and directional~\citep{giordani2018tutorial}, while the governing thresholds are fixed per deployment and adapt to no one. Learned mobility management is the response the field has converged on~\citep{lima2023deep,  yajnanarayana20205g, chien2026federated, survey2024ho}, and is now being standardized rather than merely proposed~\citep{3gpp38843,polese2023understanding}.
  
Most current solutions in literature are expensive, and some are slow. They are deep sequence models %that 
replacing a threshold comparison with a forward pass: recurrent architectures process a window step by step, attention scales quadratically, and both are dense-matrix workloads on a device that was previously comparing two numbers. Neither the energy nor the delay is reported. Efficient time-series classification does have strong cheap baselines~\cite{dempster2020rocket, dempster2021minirocket, lubba2019catch22}, but benchmarks them on accuracy and throughput~\cite{ismail2019deep}. Green AI has the measurement methodology~\cite{schwartz2020green, henderson2020towards,  patterson2021carbon} and has shown that at fleet scale inference dominates lifetime footprint~\cite{wu2022sustainable}, but has been applied almost entirely to NLP and vision, where inference is server-side, batched, and under no deadline. Wireless deployments differ on both counts~\cite{saffar2026domain}, and in our settings the constraints are explicit. The first is a deadline: measurements arrive every 40\,ms and a prediction is useful only if ready before the network acts, so a late prediction loses the benefit it was computed for while its energy is already spent. The second is energy: inference runs on the UE once per period for as long as the device is connected and moving, and at 25 inferences per second a model costing millijoules per inference would drain a smartphone within hours. Energy degrades gradually but is spent continuously and committed the moment an architecture is chosen. Together the two define a feasible region every deployable model must sit inside, and neither appears in the quantity these models are selected on.
  
In this paper we study whether a handover will trigger within the next few hundred milliseconds given a window of RSRP measurements. The task has one unusual property: its labelling function is known in closed form, so we ask learned models to recover a rule already written down. We contribute (i) an evaluation protocol that treats the deadline as a tail feasibility filter and ranks survivors on measured energy, including the feature computation usually omitted; (ii) a benchmark of eight architectures under it against what accuracy alone would select; and (iii) ablations separating feature representation from model class. We find that gradient-boosted trees dominate the accuracy-energy-latency frontier, that the size of their energy advantage depends on how inference is measured, narrowing from 10.4$\times$ under a tight loop to 3.3$\times$ under a realistic one, and no neural compression recovers the gap.

\section{Related Work}

\paragraph{Energy reporting in ML:}
  Early work on ML energy cost focused on training, estimating the energy to
  produce a single large model and its carbon
  emissions~\citep{strubell2019energy, patterson2021carbon}. This broadened into
  systematic accounting across experiments and the argument that efficiency
  belongs alongside accuracy as a criterion for
  progress~\citep{henderson2020towards, schwartz2020green}. Attention then
  shifted to deployment: at fleet scale inference can dominate a model's
  lifetime footprint~\citep{wu2022sustainable}, motivating inference energy as a
  distinct metric~\citep{desislavov2023trends}, with standardized measurement
  emerging only recently~\citep{tschand2025mlperf}. These works, however, target
  natural-language and vision workloads. The measurement techniques transfer,
  but the benchmarks do not: time-series classifiers have different input
  structures, computation patterns, and deployment requirements, making it hard
  to compare models on the efficiency axis that should drive their selection for
  energy-constrained use.
  
  \paragraph{Efficient time-series classification:}
  Before deep models became the default, the accuracy-per-cost frontier was held
  by random convolutional features such as ROCKET~\cite{dempster2020rocket,
  dempster2021minirocket} and compact interpretable feature
  sets~\cite{lubba2019catch22}. Ensemble and dictionary methods like HIVE-COTE
  pushed accuracy higher at substantially greater cost~\cite{lines2016hive,
  middlehurst2021hive}. Both random-feature and interpretable methods remain
  competitive and cheap, yet they are evaluated on accuracy and throughput, and
  reviews of the deep models report accuracy without
  energy~\cite{ismail2019deep, mohammadi2024deep}.
  
 \paragraph{Compression, trees, and the handover application:}
 Time-series compression via pruning~\cite{pruning2023ts, han2015learning},
  quantization~\cite{quant2023ts, jacob2018quantization}, and
  distillation~\citep{hinton2015distilling, gou2021knowledge} optimizes
  hardware-agnostic proxies such as parameter count or FLOPs that correlate
  only loosely with measured device energy~\cite{energy2023flops,
  cai2020tinytl}. Handover prediction itself has been tackled with
  recurrent~\cite{molina2022ho}, reinforcement-learning~\cite{wang2021rl_ho},
  and federated~\cite{zhang2023fedho} models, all evaluated on accuracy
  alone~\cite{survey2024ho}. Meanwhile tree-based models still outperform deep
  learning on tabular data~\cite{grinsztajn2022tree, shwartz2022tabular}, a
  gap usually framed through accuracy and expected to narrow on sequential
  inputs. Yet the comparison rarely accounts for energy and latency, where a
  model cheaper to both train and run accumulates its advantage across the
  deployment horizon~\cite{wu2022sustainable}. This is exactly the regime
  accuracy-only benchmarks render invisible, and we treat the handover task as
  an instrument for studying it rather than as a contribution to the
  mobility-management line.

%Prior work on ML-based HO prediction has focused on accuracy using LSTMs~\cite{molina2022ho}, reinforcement learning~\cite{wang2021rl_ho}, and federated approaches~\cite{zhang2023fedho}, but rarely considers inference cost. Recent surveys~\cite{survey2024ho} highlight the gap between academic models and deployable solutions.

%Model compression for time-series has explored pruning~\cite{pruning2023ts}, quantization~\cite{quant2023ts}, and knowledge distillation~\cite{hinton2015distilling}. However, energy measurement is typically absent or approximated with FLOPs, which correlates poorly with actual device energy~\cite{energy2023flops}. Recent work has established that gradient-boosted trees consistently
  % outperform deep learning on tabular and structured
  % data~\cite{grinsztajn2022tree}. Our contribution is not this finding per se,
  % but rather its empirical demonstration in the specific context of
  % energy-constrained cellular time-series prediction—where the practical
  % implication (deploy LightGBM at 20× lower energy) has direct engineering
  % consequence for UE-side inference.

%The Green AI movement~\cite{schwartz2020green} has motivated energy reporting, but benchmarks focus on NLP/vision tasks. To our knowledge, no prior work provides hardware-measured energy benchmarks for cellular time-series models.
\section{Problem Formulation}
We formalize the prediction task, the two constraints a deployed model
must satisfy, and the selection problem they define. The last of these is what
the rest of the paper measures.
 
\subsection{Prediction task}
We consider the problem of energy-efficient handover prediction in cellular networks. Let $\mathbf{X}_t \in \mathbb{R}^{T \times C}$ be a sliding observation window of RSRP measurements over the $T=8$ most recent timesteps, sampled at period $\Delta t = 40$\,ms, across the $C=21$ cells visible to the UE. We index cells by $c$, writing $s$ for the serving cell and $n \neq s$ for a neighbour, so that entry $(u,c)$ of $\mathbf{X}_t$ is $\mathrm{RSRP}_c(u)$, the RSRP of cell $c$ at step $u$ of the window ending at $t$. Given $\mathbf{X}_t$, the objective is to predict whether the UE will satisfy the 3GPP A3 handover condition within a future horizon of $H=8$ timesteps:
\begin{equation} 
\label{eq1}
\hat{y}_t = f_\theta(\mathbf{X}_t) \in \{0,1\},
%f_\theta(\mathbf{X}_t) \rightarrow y_t \in \{0,1\}, 
\end{equation} 
where the target label is defined explicitly as
\begin{equation}
\label{eq2}
y_t = \begin{cases}
1 & \text{if } \exists\, t_{ho} \in [t,\, t+H] \text{ at which an A3 handover fires,}\\
0 & \text{otherwise.}
\end{cases}
\end{equation}
A handover fires at $t_{ho}$ when a neighboring cell $n$ maintains a sufficiently higher signal quality than the serving cell $s$ continuously over the TTT window, that is%, when for every step index $u \in [t_{ho}-\mathrm{TTT}+1,\, t_{ho}]$
\begin{equation}
\label{eq3}
\mathrm{RSRP}_n(\illy{u}) - \mathrm{Hys} > \mathrm{RSRP}_s(\illy{u}) + \mathrm{Off},
\quad \forall \illy{u} \in [t_{ho}-\mathrm{TTT}+1,\, t_{ho}],
\end{equation}
with $\mathrm{Hys}=1$\,dB, $\mathrm{Off}=2$\,dB, and $\mathrm{TTT}=8$ steps
($320$\,ms). Labels follow the 3GPP %TS 38.331 
A3%criterion
~\citep{3gpp38331}.

\paragraph{The horizon equals the trigger window:}
By construction $H = \mathrm{TTT} = 8$ steps. Since the input window ends before the prediction instant and a positive label spans the $H$ steps up to a handover, most positive samples never observe the A3 condition complete. The model must forecast that an emerging trend will persist long enough to trigger, not read a finished trigger off its input; only the sample at the firing instant approaches detection. We set $H = \mathrm{TTT}$ deliberately: a shorter horizon would overlap more of the trigger and collapse the task toward detection, while a longer one would make the output too late to act on.

\subsection{Deployment constraints \& model selection} 
A deployed model inherits two constraints from running on the UE. They are
of different kinds, and the difference determines how each enters model
selection.

\paragraph{Timing:}
A prediction is actionable only if the action it enables can begin before the A3 timer expires. Since the pipeline must in any case keep pace with the 40\,ms measurement cadence, inference must complete within one period:
\begin{equation}
\label{eq4}
    L_\theta = L_\text{pre}(\theta) + L_\text{fwd}(\theta) + L_\text{post}(\theta)
    \;\leq\; \Delta t - (L_\text{acq} + L_\text{act}) \;\triangleq\; L_\text{max}
    \;\leq\; \Delta t = 40\,\text{ms}.
\end{equation}
The protocol terms $L_\text{acq}, L_\text{act}$ are model-independent, so we screen against the upper bound $L_\text{max}=\Delta t$; feature   preparation $L_\text{pre}$ stays inside $L_\theta$ because it is model-dependent and can rival the forward pass. A missed deadline does not stop the inference; the network simply falls back to the reactive rule, forfeiting the benefit while still paying the energy. We therefore screen at the tail: a model whose $L^{p99}\theta$ exceeds $L\text{max}$ is penalized once in a hundred inferences, or once every four seconds at 25\,Hz.

\paragraph{Energy:} 
  Inference runs once per measurement period for as long as the device is
  connected and moving, so per-inference energy accumulates against a fixed
  battery. Over a session of duration $T_S$,
  \begin{equation}
  \label{eq5}
      E_\theta \cdot \frac{T_S}{\Delta t} \;\leq\; E_\text{budget},
  \end{equation}
  where $E_\text{budget}$ is the share of the device budget allocated to this
  function and $E_\theta$ is the full-pipeline energy of one inference.
  Crucially, the processor is not busy for the whole period: it is active
  for $L_\theta$ and idle for the remaining $\Delta t - L_\theta$, a duty
  cycle of $d_\theta = L_\theta/\Delta t$, which for the models here lies
  between 0.6\% and 4.6\%. A benchmark that keeps the core continuously busy
  reports only the active-phase energy $E^{\text{busy}}_\theta$, whereas a
  duty-cycled deployment also pays for the idle intervals between inferences,
  so $E_\theta \geq E^{\text{busy}}_\theta$. In our experiments, we report both
  $E^{\text{busy}}_\theta$
  (busy-loop) and $E_\theta$ (duty-cycled), and write the budget
  constraint of Eq.~\ref{eq5} on $E_\theta$, since a device deployed at duty cycle
  $d_\theta$ incurs the full per-period cost. At 25 inferences per second,
  1\,mJ per inference is a 25\,mW continuous draw, which is substantial
  against
  a 15\,Wh battery. Unlike the timing bound, this one degrades gracefully:
  exceeding it shortens battery life rather than breaking the function. It is
  nonetheless committed at the moment the architecture is chosen, and it is
  paid once per device for the life of the deployment.

% At 25 inferences per second, 1\,mJ per inference is a 25\,mW continuous draw, which is substantial against a 15\,Wh battery. Unlike the timing bound,  this one degrades gracefully: exceeding it shortens battery life rather than breaking the function. It is nonetheless committed at the moment the architecture is chosen, and it is paid once per device for the life of the deployment.

\paragraph{Constrained model selection:}
The two constraints do not enter symmetrically. Timing is a feasibility condition, so it partitions the candidate set. Energy is a cost, so it orders what remains. For a family $\mathcal{F}$ of candidate models, define the feasible set
\begin{equation}
    \mathcal{F}_\text{feas} = \left\{ f_\theta \in \mathcal{F} \;:\;
    L^{p99}_\theta \leq L_\text{max} \right\},
\end{equation}
and select
\begin{equation}
    f^{\star} = \arg\max_{f_\theta \in \mathcal{F}_\text{feas}}
    \; \mathrm{F1}(f_\theta) \quad \text{s.t.} \quad
    E_\theta \leq E_\text{max},
\end{equation}
with $E_\text{max} = E_\text{budget}\,\Delta t / T_S$. Standard practice takes $\mathcal{F}_\text{feas} = \mathcal{F}$ and $E_\text{max} = \infty$, which recovers accuracy-only selection. The question this paper asks is what changes when they are \illy{finite}.

\begin{table}[!t]
    \begin{threeparttable}
    \centering
    \caption{Main benchmark on the 100\,s dataset (five seeds, mean
  $\pm$ std).
    Latency and energy are full-pipeline (preprocess + forward +
  postprocess).
    Energy is per-inference; \emph{Looped} keeps the core busy
  (throughput
    benchmark), \emph{Duty} runs inferences in bursts with idle
  pauses as
    deployed, and the ratio quantifies how much looped measurement
  understates
    deployment energy.
    Best neural in \textit{italics}, overall best in
  \textbf{bold}.}
    \label{tab:main_results}
    \small
    \setlength{\tabcolsep}{4pt}
    \begin{tabular}{@{}lccccccccc@{}}
    \toprule
    & & & & \multicolumn{3}{c}{Latency (ms)} &
  \multicolumn{3}{c}{Energy ($\mu$J)}
    \\
    \cmidrule(lr){5-7}\cmidrule(lr){8-10}
    Model & Params & F1 & AUC & p50 & p99 & Total$_{p50}$ & Looped
  & Duty & L/D \\
    \midrule
    \textbf{LightGBM}   & 16,317 & \textbf{0.974}$\pm$0.002 &
  \textbf{0.998} &
    \textbf{0.250} & \textbf{0.381} & 0.263 & 1,473  &
  \textbf{4,947}  & 0.30 \\
    \textit{CNN-1D}     & 81,217 & \textit{0.914}$\pm$0.012 & 0.985
  & 0.947 &
    1.095 & 0.892 & 15,316 & 16,181 & 0.95 \\
    Efficient-HO-v2     & 16,753 & 0.899$\pm$0.007 & 0.984 & 1.252
  & 1.388 & 1.372
    & 19,940 & 20,285 & 0.98 \\
    MiniRocket          & 9{,}996 & 0.872$\pm$0.011 & 0.972 & 1.321
  & 1.472 & 1.329
    & 39,596 & 44,611 & 0.89 \\
    Transformer         & 70,465 & 0.870$\pm$0.019 & 0.973 & 0.770
  & 0.899 & 0.833
    & 10,643 & 14,635 & 0.73 \\
    LSTM                & 57,665 & 0.785$\pm$0.027 & 0.951 & 0.654
  & 0.725 & 0.683
    & 9,483  & 11,723 & 0.81 \\
    Efficient-HO        & 7,618  & 0.537$\pm$0.063 & 0.848 & 0.653
  & 0.755 & 0.752
    & 8,825  & 11,069 & 0.80 \\
    catch22+ridge       & 462    & 0.371$\pm$0.000 & 0.772 & 1.838
  & 1.976 & 1.700
    & 941    & 46,883 & 0.02 \\
    \bottomrule
    \end{tabular}
    
    \footnotesize
    $^{\dagger}$ Implied active power is 0.5 W, an order of magnitude below every other model; this looped run was too too short to resolve at 50\,ms sampling, so this value (estimated as an error) is not used in any comparison.
    \end{threeparttable}
    \end{table}

% \paragraph{Task:} Given an RSRP time-series matrix $\mathbf{X} \in \mathbb{R}^{T \times C}$ where $T=8$ timesteps (320\,ms at 40\,ms period) and $C=21$ cells, predict binary label $y \in \{0, 1\}$ indicating whether a 3GPP A3 handover event will trigger within the next 320\,ms (prediction horizon $H=8$ steps).

% \paragraph{A3 Event:} Following 3GPP TS 38.331, a handover triggers when a neighbor cell $n$ satisfies:
% \begin{equation}
%     \text{RSRP}_n - \text{Hys} > \text{RSRP}_s + \text{Off}
% \end{equation}
% continuously for Time-to-Trigger (TTT) duration, where $s$ is the serving cell, Hys=1\,dB, Off=2\,dB, and TTT=320\,ms.

% \paragraph{Energy Constraint:} For a measurement period $\Delta t = 40$\,ms, inference energy $E_\text{inf}$ must satisfy:
% \begin{equation}
%     E_\text{inf} \ll E_\text{budget} / (T_\text{session} / \Delta t)
% \end{equation}
% where $E_\text{budget}$ is the UE energy budget per session. At 25 inferences/second, even 1\,mJ/inference translates to 25\,mW continuous draw which is significant for a device with a 15\,Wh battery.

\section{Experimental Setup}
\subsection{Data and models}
We generate a ray-traced RSRP dataset with Sionna~\citep{sionna2022} and SUMO vehicular mobility over a 3D model of Santa Clara, CA: 7 base station sites, 21 cells, 100 UEs over 100\,s at a 40\,ms cadence, yielding 4.7M measurements and 596 handover events. Sliding-window extraction with the labelling rule of Eqs.~\ref{eq2}-\ref{eq3} gives the classification set, balanced to a $\sim$30\% positive rate and split 70/15/15 with stratification. All results are the mean over five seeds, which control initialization, dropout and shuffling; the partition is fixed across seeds. Details are in Appendix~\ref{app:data}.

We compare three neural baselines representative of what the handover literature has converged on, three feature-based models, and two capacity probes (Table~\ref{tab:models}). LightGBM receives 247 handcrafted features, of which the min-margin and temporal-trend groups encode the A3 quantity of Eq.~\ref{eq3} almost directly; MiniRocket and catch22 are domain-agnostic cheap baselines that separate whether any advantage of the feature route comes from A3-informed features specifically or from cheap representations in general. The two Efficient-HO variants are not proposed as deployable architectures: they locate the low end of the capacity range and so test whether the neural route is recoverable by compression alone. Architectures, feature definitions and training configuration are in Appendix~\ref{app:arch}.
 
\subsection{Inference cost measurement}
Inference cost is measured on a controlled server-class CPU (36-core Xeon) under isolated load. The UE performs this inference on its application processor rather than on an accelerator, so a CPU measurement preserves the characteristic that matters while giving an isolated environment with a stable idle floor; absolute values will differ on handset silicon. Two choices distinguish the protocol from throughput benchmarking.
 
\paragraph{Full-pipeline decomposition:}
Each inference is decomposed into preprocess (normalization for the neural models, feature computation for the feature-based ones), forward, and postprocess, matching $L_\theta$ in Eq.~\ref{eq4}. All reported latency and energy cover the full pipeline. This matters because the literature times the forward pass alone, which for feature-based models omits the dominant cost.
 
\paragraph{Busy-loop versus duty-cycled energy:}
We report energy under two execution patterns. \emph{Busy-loop} runs inferences back to back with the core continuously active, which is what throughput benchmarks measure. \emph{Duty-cycled} runs inferences in short bursts separated by idle pauses (10 bursts of 50 inferences, 10\,ms pauses), so the measured average power includes the idle intervals a deployed device spends between predictions; per-inference energy is the total energy over the burst schedule divided by the number of inferences. The realized active duty cycle $d_\theta = L_\theta/\Delta t$ ranges from 0.6\% (LightGBM) to 4.6\% (catch22+ridge). Power is sampled from CPU utilization at 50\,ms resolution, so absolute $\mu$J are estimates while relative comparisons on one platform hold.
 
\subsection{Ablation protocols}
To separate representation from model class we train LightGBM and a two-layer MLP on the 247 engineered features, and LightGBM and CNN-1D on the raw flattened window, on the same split and seeds. To isolate capacity from architecture we sweep Efficient-HO-v2 across seven channel widths from 16 to 128. Knowledge distillation uses CNN-1D as teacher and the 7,618-parameter Efficient-HO as student, with temperature $T_\text{KD}=4$ and hard-label weight $\alpha=0.3$, both trained on the same split so that any gain is attributable to distillation rather than to data volume. Finally, we compare standard A3 handover against LSTM-assisted early execution, in which the handover fires as soon as the A3 condition holds in the predicted future, saving the residual TTT (Appendix~\ref{app:ablation}).

%PyTorch($\text{lr}=10^{-3}$), BCEWithLogitsLoss, batch size 64,ReduceLROnPlateau scheduler (patience=5, factor=0.5), early stopping (patience=10), max 80 epochs. LightGBM: 500 rounds, num\_leaves=63, learning rate=0.05, early stopping patience=15.
% \begin{table}[!t]
% \centering
% \caption{Main benchmark on the 100\,s dataset (five seeds, mean $\pm$ std). Latency and energy are full-pipeline (preprocess + forward + postprocess). Energy is per-inference under the realistic duty-cycled pattern. Best neural in \textit{italics}, overall best in \textbf{bold}.}
% \label{tab:main_results}
% \small
% \begin{tabular}{@{}lcccccc@{}}
% \toprule
% Model & Params & F1 & AUC & Lat.\ p50 (ms) & Lat.\ p99 (ms) & Energy ($\mu$J) \\
% \midrule
% \textbf{LightGBM}   & 16,317 & \textbf{0.974}$\pm$0.002 & \textbf{0.998} & \textbf{0.250} & \textbf{0.381} & \textbf{4,947} \\
% \textit{CNN-1D}     & 81,217 & \textit{0.914}$\pm$0.012 & 0.985 & 0.947 & 1.095 & 16,181 \\
% Efficient-HO-v2     & 16,753 & 0.899$\pm$0.007 & 0.984 & 1.252 & 1.388 & 20,285 \\
% MiniRocket          & 9{,}996 & 0.873$\pm$0.011 & 0.972 & 1.232 & 1.445 & 15{,}600 \\
% Transformer         & 70,465 & 0.870$\pm$0.019 & 0.973 & 0.770 & 0.899 & 14,635 \\
% LSTM                & 57,665 & 0.785$\pm$0.027 & 0.951 & 0.654 & 0.725 & 11,723 \\
% Efficient-HO        & 7,618  & 0.537$\pm$0.063 & 0.848 & 0.653 & 0.755 & 11,069 \\
% catch22+ridge       & 462    & 0.371$\pm$0.000 & 0.772 & 0.445 & 0.762 & 4,900 \\
% \bottomrule
% \end{tabular}
% \end{table}

\begin{figure}[!t]
\centering

% --- Left: figure ---
\begin{minipage}[t]{0.45\linewidth}
\vspace{0pt}
\centering
\includegraphics[width=\linewidth]{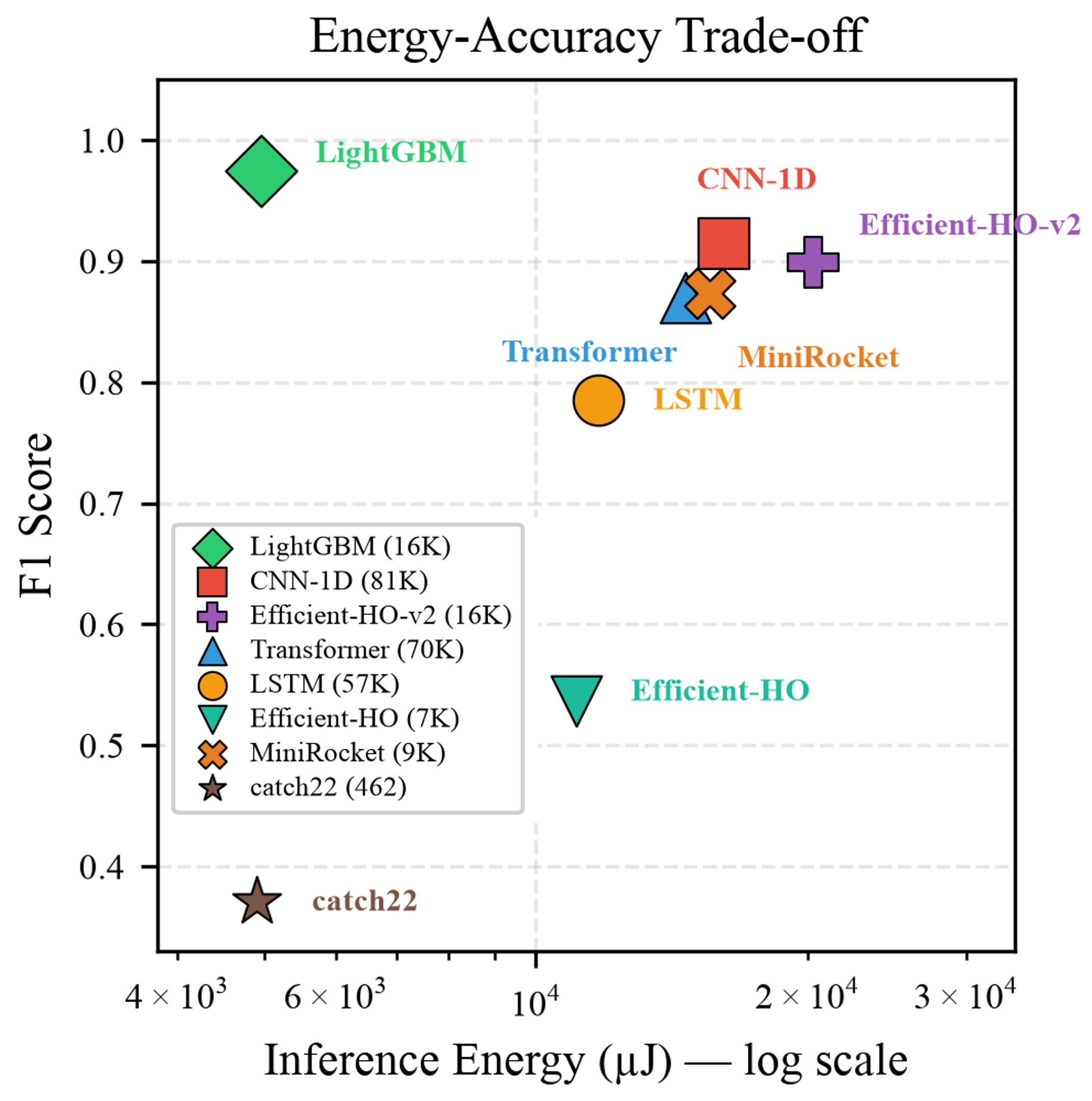}
\caption{Energy-accuracy trade-off. LightGBM dominates all neural models
in both F1 and inference energy.}
\label{fig:pareto}
\end{minipage}
\hfill
% --- Right: two stacked tables ---
\begin{minipage}[t]{0.46\linewidth}
\vspace{0pt}
\centering
\scriptsize

% Capacity sweep table
\refstepcounter{table}
\textbf{Table~\thetable:} Capacity sweep: Efficient-HO-v2 at increasing width
(five seeds).

\vspace{2pt}

\begin{tabular*}{\linewidth}{@{\extracolsep{\fill}}lrrr@{}}
\toprule
Ch. & Params & F1 & AUC \\
\midrule
16  & 1,921  & .814$\pm$.018 & .950 \\
24  & 3,433  & .840$\pm$.013 & .962 \\
32  & 5,329  & .864$\pm$.005 & .971 \\
48  & 10,273 & .893$\pm$.011 & .980 \\
64  & 16,753 & .900$\pm$.004 & .983 \\
96  & 34,321 & .911$\pm$.013 & .986 \\
128 & 58,033 & .924$\pm$.005 & .989 \\
\midrule
\textit{LGBM} & \textit{16,317} &
\textit{.974$\pm$.002} & \textit{.998} \\
\bottomrule
\end{tabular*}
\label{tab:capacity}

\vspace{8pt}

% Cross-feature ablation table
\refstepcounter{table}
\textbf{Table~\thetable:} Cross-feature ablation (five seeds).
Features vs. architecture.

\vspace{2pt}

\begin{tabular*}{\linewidth}{@{\extracolsep{\fill}}llcc@{}}
\toprule
Model & Representation & F1 & Lat(ms) \\
\midrule
LightGBM & 247 engineered    & 0.974$\pm$0.002 & 0.16 \\
LightGBM & 168 raw flattened & 0.971$\pm$0.001 & 0.15 \\
MLP      & 247 engineered    & 0.929$\pm$0.006 & 0.05 \\
CNN-1D   & raw $(8,21)$      & 0.919$\pm$0.003 & 1.50 \\
\bottomrule
\end{tabular*}
\label{tab:crossfeat}

\end{minipage}

\end{figure}
\section{Results}
\subsection{Energy-Accuracy Benchmark}
All models clear the 40\,ms deadline comfortably: even their p99 latency is at least 20$\times$ below it, so every model is feasible ($\mathcal{F}_\text{feas} = \mathcal{F}$) and selection comes down to energy. The margin also shows how much slower hardware each model could still run on. The slowest, catch22+ridge ($L^{p99}=1.98$\,ms), stays feasible until the chip is $\sim$20$\times$ slower than ours; LightGBM ($L^{p99}=0.38$\,ms) tolerates 105$\times$.

\paragraph{Finding 1 - The tree model dominates the full frontier:} LightGBM is the most accurate model (F1=0.974), the lowest-latency one (0.381\,ms p99), and the cheapest to run under the duty-cycled pattern that reflects deployment (4,947 \,$\mu$J, 2.2x below the next cheapest of the eight, Table 1, Fig. 1). Its handcrafted features (per-cell statistics, temporal trends, min-margin to neighbours) encode the A3 decision boundary directly, and its feature computation is cheap enough that the full pipeline stays under 0.4\,ms at p99. The lower busy-loop figure for catch22+ridge is an instrumentation
artefact (Table~\ref{tab:main_results}, footnote) and is excluded from the ranking. %Busy-loop measurement ranks catch22+ridge marginally lower (941 \,$\mu$J ), but that model reaches only F1=0.371 and its duty-cycled cost is an order of magnitude higher.
%LightGBM is the most accurate model (F1=0.974) and the lowest-latency one (0.381\,ms p99), and it has the lowest inference energy on both axes where a comparison is possible (Table~\ref{tab:main_results}, Fig.~\ref{fig:pareto}): 1,473 µJ busy-loop, 3.3$\times$ below the next cheapest of all eight models, and 4,947 µJ duty-cycled, 2.2$\times$ below the next cheapest of the six measured under that protocol. Its handcrafted features (per-cell statistics, temporal trends, min-margin to neighbours) encode the A3 decision boundary directly, and its feature computation is cheap enough that the full pipeline stays under 0.4\,ms at p99. MiniRocket and catch22 were measured busy-loop only; their duty-cycled energy is not reported, though their busy-loop figures already exceed LightGBM's.
%MiniRocket and catch22 were measured busy-loop only, so their energy is not comparable on this axis. Its handcrafted features (per-cell statistics, temporal trends, min-margin to neighbours) encode the A3 decision boundary directly, and its feature computation is cheap enough that the full pipeline stays under 0.4\,ms at p99. 
% \begin{figure}[h]
%   \centering
%   \includegraphics[width=0.6\linewidth]{fig/fig1_energy_accuracy_pareto.png}
%   \caption{Energy-accuracy Pareto frontier. LightGBM dominates all neural models
%   in both F1 and inference energy.}
%   \label{fig:pareto}
%   \end{figure}
\paragraph{Finding 2 - The ranking an accuracy benchmark would report is not the deployment ranking:}
CNN-1D is the best neural model on accuracy (F1=0.914) but costs over 3$\times$ the per-inference energy of LightGBM under the duty-cycled pattern and misses on accuracy as well. Every neural model is dominated. The strong cheap baseline MiniRocket reaches F1=0.872 but its 10k-kernel transform makes its full-pipeline latency 1.32\,ms, comparable to the deep models; catch22 collapses to F1=0.371, showing that cheap representations in general do not suffice and that the specific structure LightGBM's features capture is what carries the task.

\paragraph{Finding 3 - The measurement pattern is itself a result:}
Table~\ref{tab:main_results} contrasts busy-loop and duty-cycled
  energy. Busy-loop measurement understates deployment energy for every model,
  but by very different factors: the ratio is 0.30 for LightGBM against
  0.95--0.98 for CNN-1D and Efficient-HO-v2. The reason is idle time. A cheap
  model finishes its inference in a small fraction of the 40\,ms period and
  spends the rest idle, so its deployed per-inference energy is dominated by
  the idle intervals that a busy-loop benchmark never charges; a compute-heavy
  model keeps its core active for most of the period, so the two measurements
  nearly coincide. The consequence runs against our own headline: measured as
  a tight loop LightGBM appears 10.4$\times$ cheaper than CNN-1D, while
  measured under a duty-cycled schedule the advantage narrows to 3.3$\times$.
  Busy-loop benchmarking therefore does not merely omit the duty-cycled idle cost; it inflates the apparent energy advantage of the cheapest model. 

\subsection{Inference Cost Decomposition}
Table~\ref{tab:main_results} splits the p50 pipeline into its three stages. For LightGBM the feature computation (preprocess, 0.116\,ms) is  44\% of total inference and nearly equals the forward pass, exactly the cost the literature omits by timing the model alone. Reporting the forward pass only would make LightGBM appear roughly twice as fast as it is. For MiniRocket the effect is extreme: its ridge forward pass is negligible but the 10k-kernel transform dominates making it expensive in practice. For the neural models the forward pass dominates and preprocessing (normalization) is negligible, so counting the full pipeline does not help them relative to the feature models.

Training is a one-off cost and is not part of the constraints studied here. At 25\,Hz a deployed model performs of order $10^{9}$ inferences per device-year, so training energy is amortized within seconds of deployment and the inference figures above dominate the lifecycle. For completeness, training follows the same ordering as inference: LightGBM trains roughly 40$\times$ faster and 37$\times$ more energy-efficiently than CNN-1D while reaching the highest validation accuracy. Full measurements are in Appendix~\ref{app:training}.

\subsection{Feature vs.\ Architecture Ablation}
We cross the two feature representations with the two model families on the same split and seeds. Two facts emerge (Table~\ref{tab:crossfeat}). First, LightGBM gains only 0.003 from the engineered features (F1=0.974 vs F1=0.971 on the raw window), so they are not what
carries it. Second, a two-layer MLP on those same features (F1=0.929) outperforms CNN-1D on the raw window (F1=0.919) despite far less architectural capacity, so for the neural models the representation matters more than the architecture. The tree's advantage therefore comes from the model class rather than from privileged access to domain features, and the comparison is not an artefact of hand-crafted features favouring one side.

\subsection{Compression Cannot Rescue the Neural Route}
Two experiments test whether a smaller or distilled neural model can enter the feasible region without losing accuracy. First, a single-architecture capacity sweep (Table~\ref{tab:capacity}) trains Efficient-HO-v2 at seven widths on the same split. Accuracy rises smoothly and monotonically with parameters but reaches only F1=0.924 at 58K parameters, still short of LightGBM's 0.974 at a fraction of the inference cost. The gap is a capacity floor, not a tuning artefact. Second, distillation from CNN-1D into the 7.6K-parameter Efficient-HO on the full 100\,s split yields \emph{no} improvement: student F1 falls from 0.438 standalone to 0.407 distilled (mean gain $-0.031 \pm 0.033$). The apparent $+0.08$ gain reported in earlier 50\,s experiments was a data-volume effect, not a distillation effect, and disappears once teacher and student see the same data. A secondary experiment shows that although no neural model is selected by the protocol, a moderately accurate LSTM used as a confidence gate inside   the existing A3 rule can save up to 542\,ms of Time-to-Trigger latency per handover at 100\% success rate (Appendix~\ref{app:ho_sim}).% success rate; details and Table are in

% \begin{table}[h]
% \centering
% \caption{Capacity sweep: Efficient-HO-v2 at increasing width (five seeds). Accuracy climbs smoothly but never reaches the tree model.}
% \label{tab:capacity}
% \small
% \begin{tabular}{@{}lccc@{}}
% \toprule
% Channels & Params & F1 & AUC \\
% \midrule
% 16  & 1,921  & 0.814$\pm$0.018 & 0.950 \\
% 24  & 3,433  & 0.840$\pm$0.013 & 0.962 \\
% 32  & 5,329  & 0.864$\pm$0.005 & 0.971 \\
% 48  & 10,273 & 0.893$\pm$0.011 & 0.980 \\
% 64  & 16,753 & 0.900$\pm$0.004 & 0.983 \\
% 96  & 34,321 & 0.911$\pm$0.013 & 0.986 \\
% 128 & 58,033 & 0.924$\pm$0.005 & 0.989 \\
% \midrule
% \textit{LightGBM} & \textit{16,317} & \textit{0.974$\pm$0.002} & \textit{0.998} \\
% \bottomrule
% \end{tabular}
% \end{table}

%\section{Discussion: Three Conditions Generalizibilities and limitations}
\section{Discussion: Generalizability and Limitations}
The result is not specific to telecommunications; it should arise whenever three conditions hold:
\begin{enumerate}
   \item \textbf{Simple or known decision boundary:} A3 is a sustained linear inequality on cell differences, directly captured by features such as min-margin-to-neighbour. Trees exploit this structure immediately, whereas neural models must learn it from raw sequences.
   \item \textbf{Short sequence:} With $T=8$ steps, temporal inductive biases are barely exercised; attention over eight positions offers little over direct feature computation.
   \item \textbf{Duty-cycled inference:} When inference runs once per period under a deadline, energy is dominated by the per-invocation cost of leaving and re-entering idle, which throughput benchmarking amortizes away, making throughput an inappropriate measure of this regime.
\end{enumerate}
Under these conditions, constraints do not merely reorder the candidates; they rather invert the accuracy-only ranking.

%The outcome here is not specific to telecommunications. It should be expected whenever three conditions hold, and each is checkable in advance.
% \begin{enumerate}
%   │   \item \textbf{The decision boundary is simple or known.} The A3 event is
%   a sustained linear inequality on cell differences. Features such as
%   min-margin-to-neighbour encode it directly, so a tree splits on the right
%   quantity immediately while a neural model must learn the structure from raw
%   sequences.
%   │   \item \textbf{The sequence is short.} With $T=8$ steps, the temporal
%   inductive bias of recurrent and attention models is barely exercised;
%   self-attention over 8 positions buys little over direct feature computation.
%   │   \item \textbf{Inference is duty-cycled under a deadline.} When inference
%   fires once per period with the core otherwise idle, energy is dominated by
%   the single forward pass plus its feature preparation, and throughput
%   benchmarking mismeasures exactly this regime.
%   │ \end{enumerate}

%Under these conditions, constraints do not merely reorder the candidates; they invert the ranking that accuracy alone produces.

%\paragraph{Limitations}
\paragraph{Limitations} Energy is estimated from CPU utilization rather than hardware counters, so %absolute
$\mu$J are estimates while relative rankings on one platform hold.   The dataset is an %single 
urban scene (Santa Clara), leaving generalization to   other environments and mobility patterns untested. MiniRocket and catch22 use   a linear ridge head; a stronger classifier might raise their accuracy, though   not their feature-transform cost. Finally, the ray-traced simulation is   realistic but limited in temporal diversity relative to field deployment. 
\section{Conclusion}

We studied model selection for time-series classification under two deployment constraints that accuracy benchmarks do not measure: a hard per-inference deadline and a cumulative energy budget. Handover prediction in cellular networks makes both explicit, since inference runs on the UE once per 40\,ms period and must complete before the network acts. Evaluating eight architectures under a protocol that screens for deadline feasibility at the tail and ranks the survivors on measured energy, we found that a gradient-boosted tree on handcrafted features is the most accurate, the fastest and the cheapest model, and that neither a capacity sweep nor distillation brings a neural model close. How inference is measured can impact model selection: counting feature computation rather than the forward pass alone nearly doubles LightGBM's apparent cost, and measuring under a duty cycle rather than a tight loop narrows its energy advantage from 10.4$\times$ to 3.3$\times$. Under the three conditions we give, this generalizes beyond telecommunications: accuracy-only benchmarking can rank first a model that cannot be deployed.

%We evaluated time-series handover prediction under the two constraints a deployed model must satisfy: a hard timing deadline at the tail and a cumulative energy budget, with inference cost measured across the full pipeline and under a realistic duty cycle. Gradient-boosted trees on handcrafted features are simultaneously the most accurate, lowest-latency and lowest-energy feasible model, while the neural architecture an accuracy-only benchmark would select is dominated on every axis and costs several times more energy once measured as it is actually deployed. Compression does not close the gap: a capacity sweep reveals a smooth floor below the tree model, and distillation on matched data yields no gain. We gave three checkable conditions under which this reversal should be expected. Accuracy-only benchmarking does not merely under-report deployment cost; it can rank first a model that cannot be deployed. We advocate constraint-aware, full-pipeline, duty-cycled evaluation as standard practice for edge-deployed time-series models. 

% ------------------------------------------------  
\bibliographystyle{unsrt}
\bibliography{bib}

\appendix
\newpage

\section{Dataset generation}
\label{app:data}
The scene is a dense urban small-cell deployment over a 3D model of Santa Clara,
CA: 7 base station sites with 3 sectors each, giving $C=21$ cells. One hundred
UEs follow SUMO vehicular traffic patterns for 100\,s, sampled every
$\Delta t = 40$\,ms. Sionna~\citep{sionna2022} ray-traces the channel and
produces L3-filtered RSRP per UE per cell, giving 2{,}487 timestamps and 4.7M
raw measurements containing 596 handover events.
 
Labels are generated by applying Eqs.~\ref{eq2}-\ref{eq3} to the full RSRP traces, not to the
windowed data, so that the trigger condition is evaluated on the same continuous
signal the network would observe. Sliding-window extraction with stride one then
yields the classification dataset, class-balanced by subsampling negatives to a
positive rate of about 30\%, and split 70/15/15 into train, validation and test
with stratification on the label.
 
RSRP values are normalized from $[-150,-40]$\,dBm to $[0,1]$ for the neural
models; the feature-based models receive the engineered statistics of
Appendix~\ref{app:features} computed on unnormalized values.
 
%% ===========================================================================
\section{Architectures and training}
\label{app:arch}
 
The models are summarized in Table~\ref{tab:models}. Notation follows our problem statement: $\mathbf{X} \in \mathbb{R}^{T\times C}$ with $T=8$ and $C=21$, and $\sigma$ is the logistic function. 

\begin{table}[!th]
\centering
\caption{Evaluated models. Parameter counts for the tree models are total leaves
across all boosting rounds; MiniRocket and catch22 use a ridge classifier, so
their count is the feature dimension.}
\label{tab:models}
\small
\begin{tabular}{@{}llll@{}}
\toprule
Model & Params & Family & Configuration \\
\midrule
LSTM            & 57,665 & recurrent    & 2 layers, hidden 64, dropout 0.2 \\
Transformer     & 70,465 & attention    & $d_\text{model}{=}64$, 2 layers, 4 heads \\
CNN-1D          & 81,217 & convolutional & 3 residual blocks, 64 channels, kernel 3 \\
\midrule
LightGBM        &  16,317 & trees + features & 500 rounds, 63 leaves, 247 features \\
MiniRocket      & 9{,}996 feat.\ & random features  & 10k kernels, ridge \\
catch22+ridge   & 462 feat.\ & feature + linear & 22 features $\times$ 21 cells, ridge \\
\midrule
Efficient-HO    & 7,618  & GRU + early exit & exit gate $\tau{=}0.9$ \\
Efficient-HO-v2 & 16,753 & DS-CNN       & 3 DS-Conv blocks,  64 channels \\
\bottomrule
\end{tabular}
\end{table}

\subsection{Neural baselines}
 
\paragraph{LSTM (57,665 parameters):}
Two stacked LSTM layers, hidden size 64, dropout 0.2 between layers, applied
along the time axis. The final hidden state feeds a head $64 \to 32 \to 1$ with
ReLU between the linear layers.
 
\paragraph{Transformer (70,465 parameters):}
A linear projection maps each timestep from $C=21$ to $d_\text{model}=64$ and a
learnable positional encoding is added. Two encoder layers follow, each with 4
attention heads, feed-forward width 128, GELU and dropout 0.1. Global average
pooling over the 8 positions precedes a linear classifier.
 
\paragraph{CNN-1D (81,217 parameters):}
An input convolution lifts 21 channels to 64, followed by three residual blocks.
Each block applies Conv1D (kernel 3, padding 1), BatchNorm, ReLU and dropout
0.1, twice, with an identity skip. Global average pooling precedes a linear
classifier.
 
\subsection{Feature-based models}
 
\paragraph{LightGBM:}
Gradient-boosted trees over the 247 features of Appendix~\ref{app:features},
with 500 boosting rounds, 63 leaves, learning rate 0.05 and early stopping
patience 15.
 
\paragraph{MiniRocket (9,996 features):}
The MiniRocket transform~\citep{dempster2021minirocket} with 10{,}000 kernels
applied per cell channel, producing proportion-of-positive-values features,
followed by a ridge classifier with regularization selected on the validation
split from $\{10^{-3},\dots,10^{3}\}$.
 
\paragraph{catch22 + ridge (462 features):}
The 22 canonical characteristics of~\citep{lubba2019catch22} computed
independently for each of the 21 cell channels, followed by a ridge classifier
selected as above. The ridge solution is deterministic on a fixed split, which
is why its standard deviation across seeds is zero in
Table~\ref{tab:main_results}.
 
\subsection{Capacity probes}
 
\paragraph{Efficient-HO (7,618 parameters):}
A linear stage maps the flattened window to a hidden representation, an
early-exit gate decides whether recurrent processing is needed, and a GRU and
classifier produce the output when it is:
\begin{align}
\mathbf{h}_1 &= \mathrm{ReLU}\!\left(\mathbf{W}_1\,\mathrm{flatten}(\mathbf{X}) + \mathbf{b}_1\right), &
p_\text{exit} &= \sigma\!\left(\mathbf{w}_e^\top \mathbf{h}_1 + b_e\right), \\
\mathbf{h}_2 &= \mathrm{GRU}\!\left(\mathrm{reshape}(\mathbf{h}_1)\right), &
\hat{y} &= \sigma\!\left(\mathbf{W}_3\,\mathrm{ReLU}\!\left(\mathbf{W}_2 \mathbf{h}_2 + \mathbf{b}_2\right) + \mathbf{b}_3\right).
\end{align}
At inference the model returns the gate decision and skips the GRU path when
$\max(p_\text{exit}, 1-p_\text{exit}) > \tau$ with $\tau = 0.9$. Training
minimizes $\mathcal{L} = \mathcal{L}_\text{main} + 0.3\,\mathcal{L}_\text{exit}$,
where $\mathcal{L}_\text{exit}$ is the binary cross-entropy of the gate against
the target.

\paragraph{Efficient-HO-v2 / DS-CNN (16,753 parameters):}
Three depthwise-separable convolution blocks~\citep{nascimento2019dsconv} with
residual connections; each applies a depthwise Conv1D (kernel 3) then a
pointwise Conv1D, BatchNorm and ReLU. Global average pooling precedes a linear
classifier, and all operations are INT8-quantizable. The capacity sweep varies
the channel width over $\{16, 24, 32, 48, 64, 96, 128\}$ with all else fixed.
 
\subsection{Training configuration}
Neural models use Adam with initial learning rate $10^{-3}$, BCEWithLogitsLoss,
batch size 64, at most 80 epochs, ReduceLROnPlateau (factor 0.5, patience 5) and
early stopping on validation loss (patience 10). Seeds
$\{42,123,456,789,1024\}$ control weight initialization, dropout masks and data
shuffling; the partition is fixed across seeds.
 
%% ===========================================================================
\section{Engineered feature set}
\label{app:features}
 
The 247 features supplied to LightGBM are, writing $m_c(u)$ for the RSRP of cell
$c$ at step $u$ of the window and $s$ for the serving cell:
 
\begin{itemize}\itemsep2pt
  \item the flattened window, $T \times C = 168$ values;
  \item per-cell mean $\bar{m}_c = \frac{1}{T}\sum_u m_c(u)$, 21 values;
  \item per-cell standard deviation over the window, 21 values;
  \item per-cell temporal trend, the slope of a least-squares fit of $m_c(u)$
        against $u$, 21 values;
  \item max-per-step, $\max_c m_c(u)$, one per timestep, 8 values;
  \item min-margin, $\min_{n \neq s}\left[m_s(u) - m_n(u)\right]$, one per
        timestep, 8 values.
\end{itemize}
 
The last group is the A3 quantity of Eq.~3 evaluated per timestep, up to the
constants $\mathrm{Hys}$ and $\mathrm{Off}$, and the trend group stands in for
its sustained-over-TTT character. This is the sense in which the representation
encodes the labelling function, and it is what the ablation of
Appendix~\ref{app:ablation} manipulates.
 
%% ===========================================================================
\section{Ablation protocols}
\label{app:ablation}
 
\paragraph{Features versus architecture:}
Two representations, the 247 engineered features and the 168-value raw flattened
window, are crossed with two model families, LightGBM and a two-layer MLP with
hidden width 64. CNN-1D on the raw window is included from the main benchmark.
All cells use the split and seeds of Appendix~\ref{app:data}.
 
\paragraph{Capacity:}
Efficient-HO-v2 is trained at seven channel widths with all other settings
fixed, isolating capacity from architecture.
 
\paragraph{Knowledge distillation:}
Teacher CNN-1D, student the 7,618-parameter Efficient-HO:
\begin{equation}
    \mathcal{L}_\text{KD} = \alpha\,\mathrm{BCE}(\hat{y}, y) +
    (1-\alpha)\,\mathrm{BCE}\!\left(\hat{y}/T_\text{KD},\,
    \sigma(z^{\text{teach}}/T_\text{KD})\right),
\end{equation}
with $T_\text{KD}=4$, $\alpha=0.3$, 100 epochs and patience 10. Teacher and
student are trained on the same split.
 
\paragraph{Early-execution handover:}
  During the TTT countdown an ONNX-exported LSTM predicts future RSRP; if the A3
  condition holds in the predicted future the handover executes immediately,
  saving the residual TTT. The $320$\,ms TTT of Eq.~3 is the labeling horizon
  the
  classifier is trained on, whereas the simulation sweeps the network's enforced
  TTT and reports the best configuration at $640$\,ms; the longer enforced timer
  is what leaves a residual for early execution to recover. The two algorithms
  act
  on different handover populations, so their counts are not a like-for-like
  comparison: the baseline (Algo~0) counts all $489$ A3 triggers, while early
  execution (Algo~4) pre-empts marginal and oscillatory triggers before they
  mature into separate events, yielding $238$ executed handovers and fewer
  ping-pongs ($5\!\to\!2$). The $542$\,ms in Table~\ref{tab:ho_sim} is therefore a per-handover average over the $223$ early-executed events, not a paired difference against the baseline count.
 
%% ===========================================================================
\section{Training cost}
\label{app:training}
 
Training energy is reported for completeness; it is not one of the
deployment constraints and does not enter the selection of Eq.~7.
Measurements use CodeCarbon on the same platform as the inference results, over
the same five seeds. The ordering matches the inference ordering: LightGBM
trains 40.6$\times$ faster and uses 37.7$\times$ less energy than CNN-1D while
reaching the highest validation accuracy. At the deployment cadence of 25\,Hz
this difference is amortized within seconds, which is why the paper treats
inference rather than training as the operative cost.
 
\begin{table}[h]
\centering
\caption{Training cost on server CPU, measured with CodeCarbon (five seeds, mean
$\pm$ std). Same platform as the inference results.}
\label{tab:training}
\small
\begin{tabular}{@{}lccccc@{}}
\toprule
Model & Params & Train Time (s) & Energy (kWh) & CO$_2$ (g) & Best Val Acc \\
\midrule
LightGBM        & 16,317 & \phantom{00}4.2$\pm$0.2\phantom{0} & $1.3\times10^{-4}$ & 0.005 & \textbf{0.981} \\
LSTM            & 57,665 & 122.1$\pm$5.5\phantom{0} & $3.5\times10^{-3}$ & 0.141 & 0.885 \\
Efficient-HO    & \phantom{0}7,618 & 134.4$\pm$5.9\phantom{0} & $3.8\times10^{-3}$ & 0.156 & 0.808 \\
Efficient-HO-v2 & 16,753 & 146.8$\pm$13.9 & $4.8\times10^{-3}$ & 0.196 & 0.942 \\
CNN-1D          & 81,217 & 170.6$\pm$10.2 & $4.9\times10^{-3}$ & 0.198 & 0.945 \\
Transformer     & 70,465 & 208.9$\pm$5.1\phantom{0} & $5.9\times10^{-3}$ & 0.242 & 0.929 \\
\bottomrule
\end{tabular}
\end{table}

\section{Handover Simulation}\label{app:ho_sim}
Although no neural model is selected by the protocol, a moderately accurate LSTM still has value as a confidence gate inside the existing rule. Table~\ref{tab:ho_sim} compares baseline A3 (Algo~0) with LSTM early execution (Algo~4). At TTT=640\,ms the LSTM saves an average of 542\,ms per handover at 100\% success rate while reducing ping-pongs, showing that prediction can accelerate the standard procedure without replacing it.

\begin{table}[h]
\centering
\caption{HO simulation: baseline A3 (Algo~0) vs.\ LSTM early execution (Algo~4) at best configuration, TTT=640\,ms.}
\label{tab:ho_sim}
\small
\begin{tabular}{@{}lcccccc@{}}
\toprule
Algorithm & Total HOs & Success & Failures & Ping-Pong & Early Exec & TTT Saved (ms) \\
\midrule
Algo 0 (baseline)   & 489 & 489 & 0 & 5 & --- & --- \\
Algo 4 (early exec) & 238 & 238 & 0 & 2 & 223 & 542 \\
\bottomrule
\end{tabular}
\end{table}

\end{document}